\documentclass[10pt,twocolumn,letterpaper]{article}

\usepackage{iccv}
\usepackage{times}
\usepackage{epsfig}
\usepackage{graphicx}
\usepackage{amsmath}
\usepackage{amssymb}

\usepackage[breaklinks=true,bookmarks=false]{hyperref}

\iccvfinalcopy 

\def\iccvPaperID{****} 

\begin{document}
\title{Learning Deep Features for Discriminative Localization}

\author{
Bolei Zhou, Aditya Khosla, Agata Lapedriza, Aude Oliva, Antonio Torralba \\
Computer Science and Artificial Intelligence Laboratory, MIT\\
\texttt{\{bzhou,khosla,agata,oliva,torralba\}@csail.mit.edu}
}

\maketitle

\begin{abstract}

In this work, we revisit the global average pooling layer proposed in \cite{lin2013network}, and shed light on how it explicitly enables the convolutional neural network to have remarkable localization ability despite being trained on image-level labels. While this technique was previously proposed as a means for regularizing training, we find that it actually builds a generic localizable deep representation that can be applied to a variety of tasks. Despite the apparent simplicity of global average pooling, we are able to achieve 37.1\% top-5 error for object localization on ILSVRC 2014, which is remarkably close to the 34.2\% top-5 error achieved by a fully supervised CNN approach. We demonstrate that our network is able to localize the discriminative image regions on a variety of tasks despite not being trained for them.

\end{abstract}

\section{Introduction}

Recent work by Zhou \textit{et al}~\cite{zhou2014object} has shown that the convolutional units of various layers of convolutional neural networks (CNNs) actually behave as object detectors despite no supervision on the location of the object was provided. Despite having this remarkable ability to localize objects in the convolutional layers, this ability is lost when fully-connected layers are used for classification. Recently some popular fully-convolutional neural networks such as the Network in Network (NIN) \cite{lin2013network} and GoogLeNet \cite{szegedy2014going} have been proposed to avoid the use of fully-connected layers to minimize the number of parameters while maintaining high performance. 

\begin{figure}
\begin{center}
\includegraphics[width=1\linewidth]{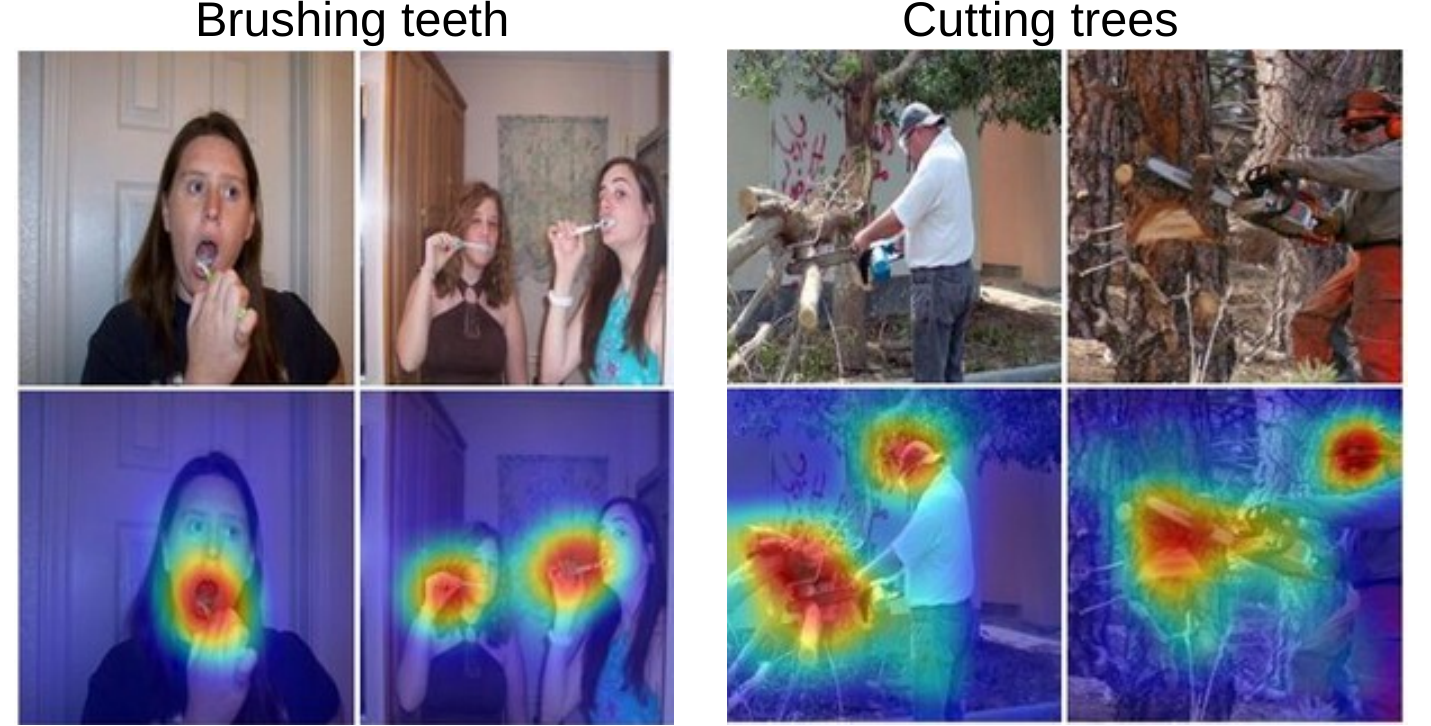}
\end{center}
\vspace*{-4mm}
\caption{A simple modification of the global average pooling layer combined with our class activation mapping (CAM) technique allows the classification-trained CNN to both classify the image and localize class-specific image regions in a single forward-pass e.g., the toothbrush for \textit{brushing teeth} and the chainsaw for \textit{cutting trees}.}\label{figure_cover}
\end{figure}

In order to achieve this, \cite{lin2013network} uses \textit{global average pooling} which acts as a structural regularizer, preventing overfitting during training. In our experiments, we found that the advantages of this global average pooling layer extend beyond simply acting as a regularizer - In fact, with a little tweaking, the network can retain its remarkable localization ability until the final layer. This tweaking allows identifying easily the discriminative image regions in a single forward-pass for a wide variety of tasks, even those that the network was not originally trained for. As shown in Figure \ref{figure_cover}(a), a CNN trained on object categorization is successfully able to localize the discriminative regions for action classification as the objects that the humans are interacting with rather than the humans themselves. 

Despite the apparent simplicity of our approach, for the weakly supervised object localization on ILSVRC benchmark~\cite{ILSVRCijcv15}, our best network achieves 37.1\% top-5 test error, which is rather close to the 34.2\% top-5 test error achieved by fully supervised AlexNet~\cite{krizhevsky2012imagenet}. Furthermore, we demonstrate that the localizability of the deep features in our approach can be easily transferred to other recognition datasets for generic classification, localization, and concept discovery.\footnote{Our models are available at: \href{http://cnnlocalization.csail.mit.edu}{http://cnnlocalization.csail.mit.edu}}.

\subsection{Related Work}
Convolutional Neural Networks (CNNs) have led to impressive performance on a variety of visual recognition tasks~\cite{krizhevsky2012imagenet,zhou2014learning,girshick2014rich}. Recent work has shown that despite being trained on image-level labels, CNNs have the remarkable ability to localize objects~\cite{bergamo2014self,oquab2014weakly,cinbis2015weakly,oquab2014learning}. In this work, we show that, using the right architecture, we can generalize this ability beyond just localizing objects, to start identifying exactly which regions of an image are being used for discrimination. Here, we discuss the two lines of work most related to this paper: weakly-supervised object localization and visualizing the internal representation of CNNs.

\textbf{Weakly-supervised object localization:} There have been a number of recent works exploring weakly-supervised object localization using CNNs~\cite{bergamo2014self,oquab2014weakly,cinbis2015weakly,oquab2014learning}. Bergamo \textit{et al}~\cite{bergamo2014self} propose a technique for self-taught object localization involving masking out image regions to identify the regions causing the maximal activations in order to localize objects. Cinbis \textit{et al}~\cite{cinbis2015weakly} combine multiple-instance learning with CNN features to localize objects.  
Oquab \textit{et al}~\cite{oquab2014learning} propose a method for transferring mid-level image representations and show that some object localization can be achieved by evaluating the output of CNNs on multiple overlapping patches. However, the authors do not actually evaluate the localization ability. On the other hand, while these approaches yield promising results, they are not trained end-to-end and require multiple forward passes of a network to localize objects, making them difficult to scale to real-world datasets. Our approach is trained end-to-end and can localize objects in a single forward pass.

The most similar approach to ours is the work based on global max pooling by Oquab \textit{et al}~\cite{oquab2014weakly}. Instead of global \textit{average} pooling, they apply global \textit{max} pooling to localize a point on objects. However, their localization is limited to a point lying in the boundary of the object rather than determining the full extent of the object. We believe that while the \textit{max} and \textit{average} functions are rather similar, the use of average pooling encourages the network to identify the complete extent of the object. The basic intuition behind this is that the loss for average pooling benefits when the network identifies \textit{all} discriminative regions of an object as compared to max pooling. This is explained in greater detail and verified experimentally in Sec.~\ref{sec:locresults}. Furthermore, unlike~\cite{oquab2014weakly}, we demonstrate that this localization ability is generic and can be observed even for problems that the network was not trained on.

We use \textit{class activation map} to refer to the weighted activation maps generated for each image, as described in Section~\ref{sec:cam}. We would like to emphasize that while global average pooling is not a novel technique that we propose here, the observation that it can be applied for accurate discriminative localization is, to the best of our knowledge, unique to our work. We believe that the simplicity of this technique makes it portable and can be applied to a variety of computer vision tasks for fast and accurate localization.

\textbf{Visualizing CNNs:} There has been a number of recent works~\cite{zeiler2014visualizing,mahendran2004understanding,dosovitskiy2015inverting,zhou2014object} that visualize the internal representation learned by CNNs in an attempt to better understand their properties. Zeiler \textit{et al}~\cite{zeiler2014visualizing} use deconvolutional networks to visualize what patterns activate each unit. Zhou \textit{et al.}~\cite{zhou2014object} show that CNNs learn object detectors while being trained to recognize scenes, and demonstrate that the same network can perform both scene recognition and object localization in a single forward-pass. Both of these works only analyze the convolutional layers, ignoring the fully-connected thereby painting an incomplete picture of the full story. By removing the fully-connected layers and retaining most of the performance, we are able to understand our network from the beginning to the end. 

Mahendran \textit{et al}~\cite{mahendran2004understanding} and Dosovitskiy \textit{et al}~\cite{dosovitskiy2015inverting} analyze the visual encoding of CNNs by inverting deep features at different layers. While these approaches can invert the fully-connected layers, they only show what information is being preserved in the deep features without highlighting the relative importance of this information. Unlike~\cite{mahendran2004understanding} and~\cite{dosovitskiy2015inverting}, our approach  can highlight exactly which regions of an image are important for discrimination. Overall, our approach provides another glimpse into the soul of CNNs.

\section{Class Activation Mapping}
\label{sec:cam}

\begin{figure*}
\includegraphics[width=1\linewidth]{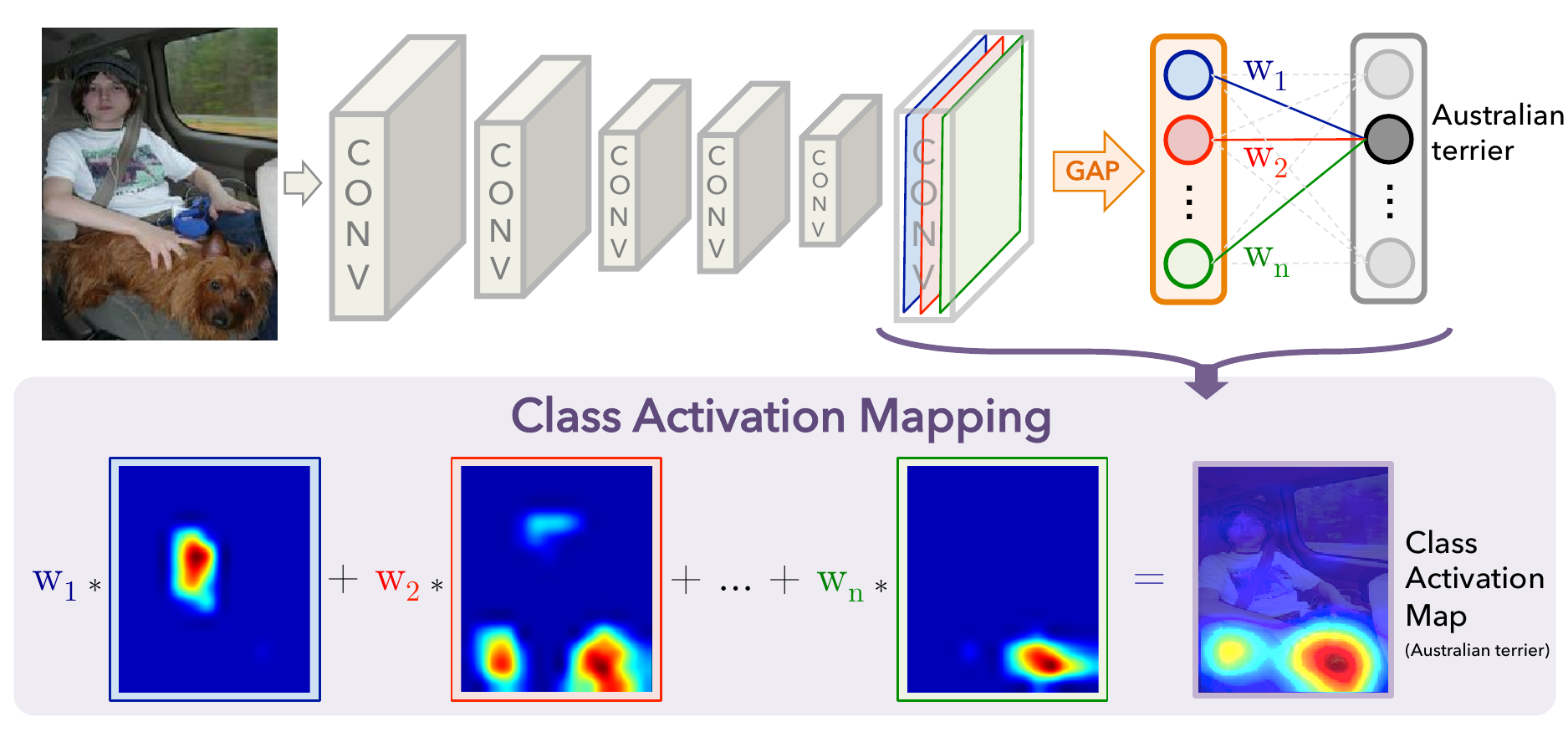}
 \caption{Class Activation Mapping: the predicted class score is mapped back to the previous convolutional layer to generate the class activation maps (CAMs). The CAM highlights the class-specific discriminative regions.}\label{figure_module}
\end{figure*}

\begin{figure}
\begin{center}
\includegraphics[width=1\linewidth]{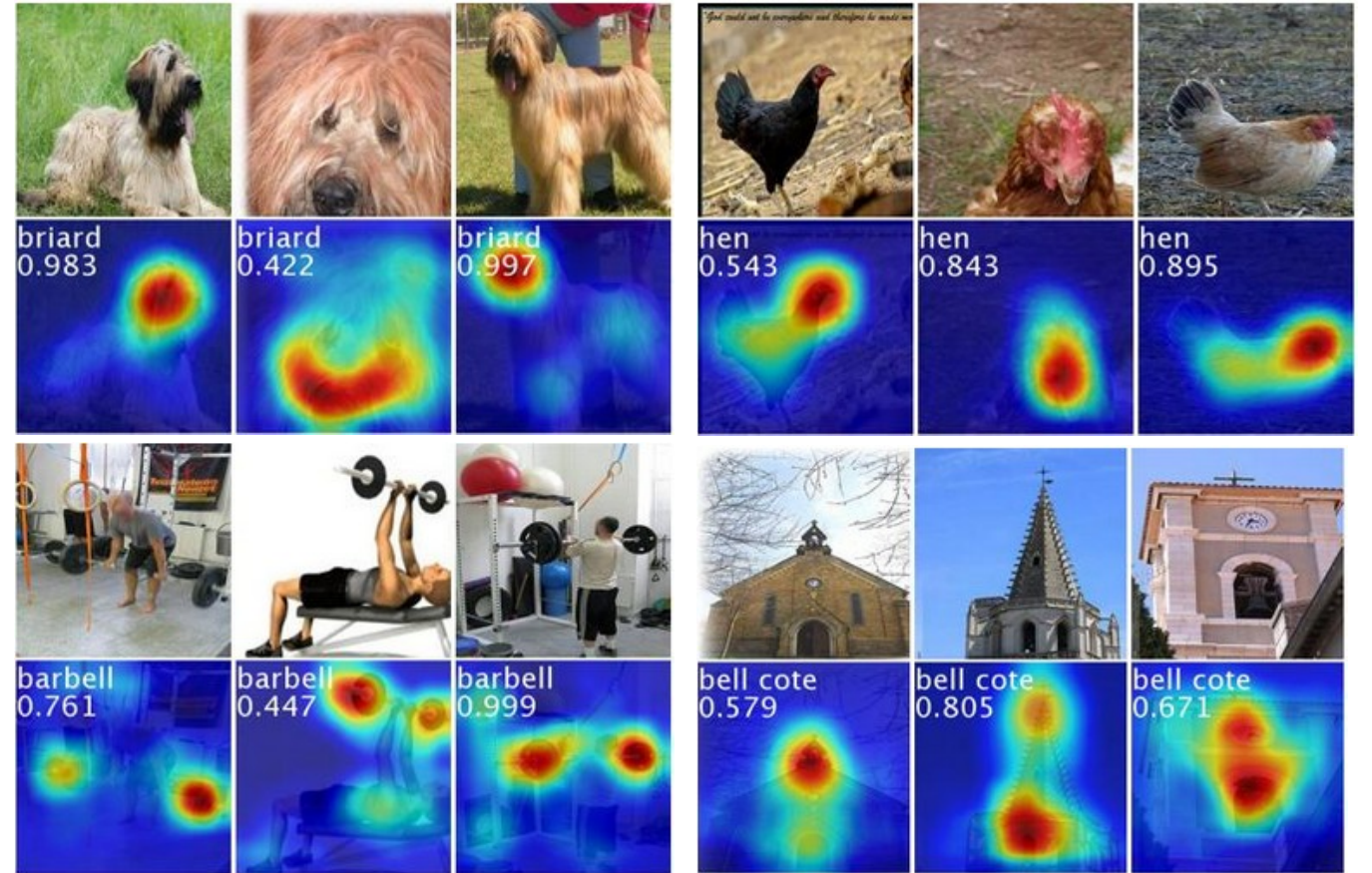}
\end{center}
\vspace*{-4mm}
 \caption{The CAMs of four classes from ILSVRC~\cite{ILSVRCijcv15}. The maps highlight the discriminative image regions used for image classification e.g., the head of the animal for \textit{briard} and \textit{hen}, the plates in \textit{barbell}, and the bell in \textit{bell cote}.}\label{figure_examplemapping}
\end{figure}

In this section, we describe the procedure for generating \textit{class activation maps} (CAM) using global average pooling (GAP) in CNNs. A class activation map for a particular category indicates the discriminative image regions used by the CNN to identify that category (e.g., Fig.~\ref{figure_examplemapping}). The procedure for generating these maps is illustrated in Fig.~\ref{figure_module}.

We use a network architecture similar to Network in Network~\cite{lin2013network} and GoogLeNet~\cite{szegedy2014going} - the network largely consists of convolutional layers, and just before the final output layer (softmax in the case of categorization), we perform global average pooling on the convolutional feature maps and use those as features for a fully-connected layer that produces the desired output (categorical or otherwise). Given this simple connectivity structure, we can identify the importance of the image regions by projecting back the weights of the output layer on to the convolutional feature maps, a technique we call class activation mapping.


As illustrated in Fig.~\ref{figure_module}, global average pooling outputs the spatial average of the feature map of each unit at the last convolutional layer. A weighted sum of these values is used to generate the final output. Similarly, we  compute a weighted sum of the feature maps of the last convolutional layer to obtain our class activation maps. We describe this more formally below for the case of softmax. The same technique can be applied to regression and other losses.


For a given image, let $f_{k}(x,y)$ represent the activation of unit $k$ in the last convolutional layer at spatial location $(x, y)$. Then, for unit $k$, the result of performing global average pooling, $F^{k}$ is $\sum_{x,y}f_{k}(x,y)$. Thus, for a given class $c$, the input to the softmax, $S_c$, is $\sum_{k}w^{c}_{k}F_{k}$ where $w^{c}_{k}$ is the weight corresponding to class $c$ for unit $k$. Essentially, $w^{c}_{k}$ indicates the \textit{importance} of $F_{k}$ for class $c$. Finally the output of the softmax for class $c$, $P_c$ is given by $\frac{\exp(S_c)}{\sum_{c}\exp(S_c)}$. Here we ignore the bias term: we explicitly set the input bias of the softmax to $0$ as it has little to no impact on the classification performance.


By plugging $F_{k} = \sum_{x,y}f_{k}(x,y)$ into the class score, $S_c$, we obtain
\begin{align}\label{eq:linearscore}
\nonumber S_c = &\sum_{k}w^{c}_{k}\sum_{x,y}f_{k}(x,y)\\
=&\sum_{x,y}\sum_{k}w^{c}_{k}f_{k}(x,y).
\end{align}
We define $M_{c}$ as the class activation map for class $c$, where each spatial element is given by
\begin{align}
M_{c}(x,y) = \sum_{k}w^{c}_{k}f_{k}(x,y).
\end{align} 
Thus, $S_c =\sum_{x,y}M_{c}(x,y)$, and hence $M_{c}(x,y)$ directly indicates the importance of the activation at spatial grid $(x,y)$ leading to the classification of an image to class $c$.

Intuitively, based on prior works~\cite{zhou2014object,zeiler2014visualizing}, we expect each unit to be activated by some visual pattern within its receptive field. Thus $f_k$ is the map of the presence of this visual pattern. The class activation map is simply a weighted linear sum of the presence of these visual patterns at different spatial locations. By simply upsampling the class activation map to the size of the input image, we can identify the image regions most relevant to the particular category.

\begin{figure}
\begin{center}
\includegraphics[width=1\linewidth]{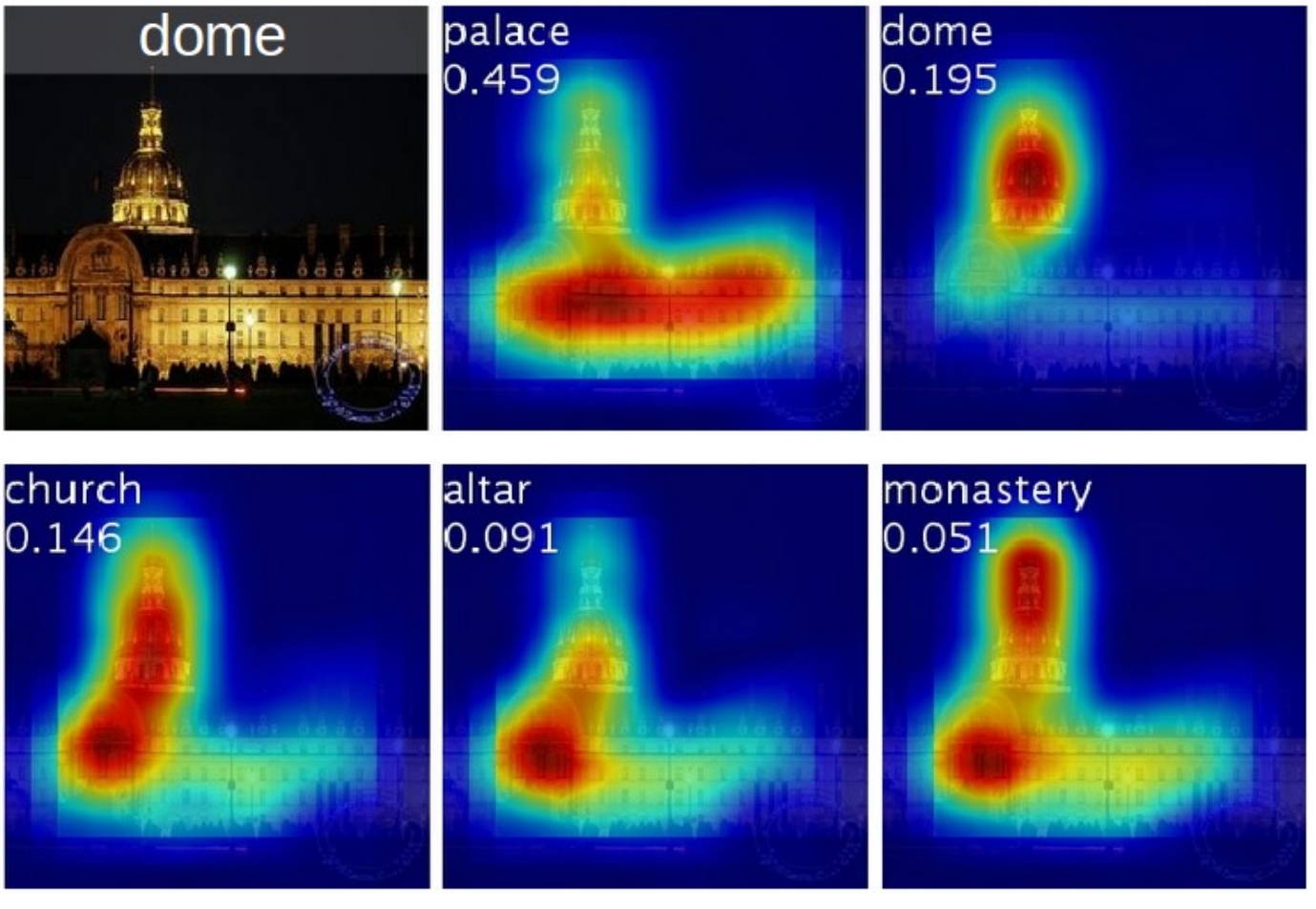}
\end{center}
\vspace*{-4mm}
 \caption{Examples of the CAMs generated from the top 5 predicted categories for the given image with ground-truth as dome. The predicted class and its score are shown above each class activation map. We observe that the highlighted regions vary across predicted classes e.g., \textit{dome} activates the upper round part while \textit{palace} activates the lower flat part of the compound.}\label{figure_multipleprediction}
\end{figure}

In Fig.~\ref{figure_examplemapping}, we show some examples of the CAMs output using the above approach. We can see that the discriminative regions of the images for various classes are highlighted. In Fig.~\ref{figure_multipleprediction} we highlight the differences in the CAMs for a single image when using different classes $c$ to generate the maps. We observe that the discriminative regions for different categories are different even for a given image. This suggests that our approach works as expected. We demonstrate this quantitatively in the sections ahead.

\textbf{Global average pooling (GAP) vs global max pooling (GMP):} Given the prior work~\cite{oquab2014weakly} on using GMP for weakly supervised object localization, we believe it is important to highlight the intuitive difference between GAP and GMP. We believe that GAP loss encourages the network to identify the extent of the object as compared to GMP which encourages it to identify just one discriminative part. This is because, when doing the average of a map, the value can be maximized by finding \textit{all} discriminative parts of an object as all low activations reduce the output of the particular map. On the other hand, for GMP, low scores for all image regions except the most discriminative one do not impact the score as you just perform a max. We verify this experimentally on ILSVRC dataset in Sec.~\ref{sec:weaklocalization}: while GMP achieves similar classification performance as GAP, GAP outperforms GMP for localization.

\section{Weakly-supervised Object Localization}
\label{sec:weaklocalization}
In this section, we evaluate the localization ability of CAM when trained on the ILSVRC 2014 benchmark dataset~\cite{ILSVRCijcv15}. We first describe the experimental setup and the various CNNs used in Sec.~\ref{sec:locsetup}. Then, in Sec.~\ref{sec:locresults} we verify that our technique does not adversely impact the classification performance when learning to localize and provide detailed results on weakly-supervised object localization.


\subsection{Setup}
\label{sec:locsetup}
For our experiments we evaluate the effect of using CAM on the following popular CNNs: AlexNet~\cite{krizhevsky2012imagenet}, VGGnet~\cite{simonyan2014very}, and GoogLeNet~\cite{szegedy2014going}. In general, for each of these networks we remove the fully-connected layers before the final output and replace them with GAP followed by a fully-connected softmax layer. 

We found that the localization ability of the networks improved when the last convolutional layer before GAP had a higher spatial resolution, which we term the \textit{mapping resolution}. In order to do this, we removed several convolutional layers from some of the networks. Specifically, we made the following modifications: For AlexNet, we removed the layers after \texttt{conv5}
(i.e., \texttt{pool5} to \texttt{prob}) resulting in a mapping resolution of $13 \times 13$. For VGGnet, we removed the layers after \texttt{conv5-3} (i.e., \texttt{pool5} to \texttt{prob}), resulting in a mapping resolution of $14 \times 14$. For GoogLeNet, we removed the layers after \texttt{inception4e} (i.e., \texttt{pool4} to \texttt{prob}), resulting in a mapping resolution of $14 \times 14$. To each of the above networks, we added a convolutional layer of size $3\times3$, stride $1$, pad $1$ with 1024 units, followed by a GAP layer and a softmax layer. Each of these networks were then fine-tuned\footnote{Training from scratch also resulted in similar performances.} on the 1.3M training images of ILSVRC~\cite{ILSVRCijcv15} for 1000-way object classification resulting in our final networks AlexNet-GAP, VGGnet-GAP and GoogLeNet-GAP respectively.

For classification, we compare our approach against the original AlexNet~\cite{krizhevsky2012imagenet}, VGGnet~\cite{simonyan2014very}, and GoogLeNet~\cite{szegedy2014going}, and also provide results for Network in Network (NIN)~\cite{lin2013network}. For localization, we compare against the original GoogLeNet\footnote{This has a lower mapping resolution than GoogLeNet-GAP.}, NIN and using backpropagation~\cite{simonyan2013deep} instead of CAMs. Further, to compare average pooling against max pooling, we also provide results for GoogLeNet trained using global max pooling (GoogLeNet-GMP).

We use the same error metrics (top-1, top-5) as ILSVRC for both classification and localization to evaluate our networks. For classification, we evaluate on the ILSVRC validation set, and for localization we evaluate on both the validation and test sets.

\subsection{Results}
\label{sec:locresults}
We first report results on object classification to demonstrate that our approach does not significantly hurt classification performance. Then we demonstrate that our approach is effective at weakly-supervised object localization.

\textbf{Classification:} Tbl.~\ref{network_classificationValidation} summarizes the classification performance of both the original and our GAP networks. We find that in most cases there is a small performance drop of $1 - 2\%$ when removing the additional layers from the various networks. We observe that AlexNet is the most affected by the removal of the fully-connected layers. To compensate, we add two convolutional layers just before GAP resulting in the AlexNet*-GAP network. We find that AlexNet*-GAP performs comparably to AlexNet. Thus, overall we find that the classification performance is largely preserved for our GAP networks. Further, we observe that GoogLeNet-GAP and GoogLeNet-GMP have similar performance on classification, as expected. Note that it is important for the networks to perform well on classification in order to achieve a high performance on localization as it involves identifying both the object category and the bounding box location accurately.

\begin{figure*}
\begin{center}
\includegraphics[width=1\textwidth]{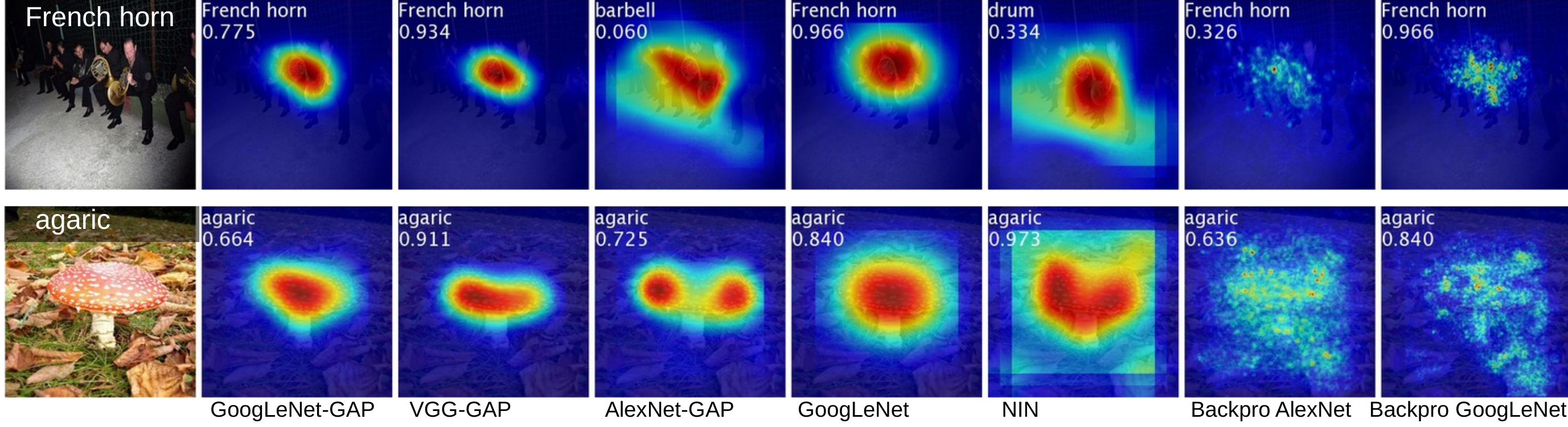}
\end{center}
\vspace{-4mm}
\caption{Class activation maps from CNN-GAPs and the class-specific saliency map from the backpropagation methods.}
\label{fig:activationmap}
\end{figure*}

\begin{figure*}
\begin{center}
\includegraphics[width=1\textwidth]{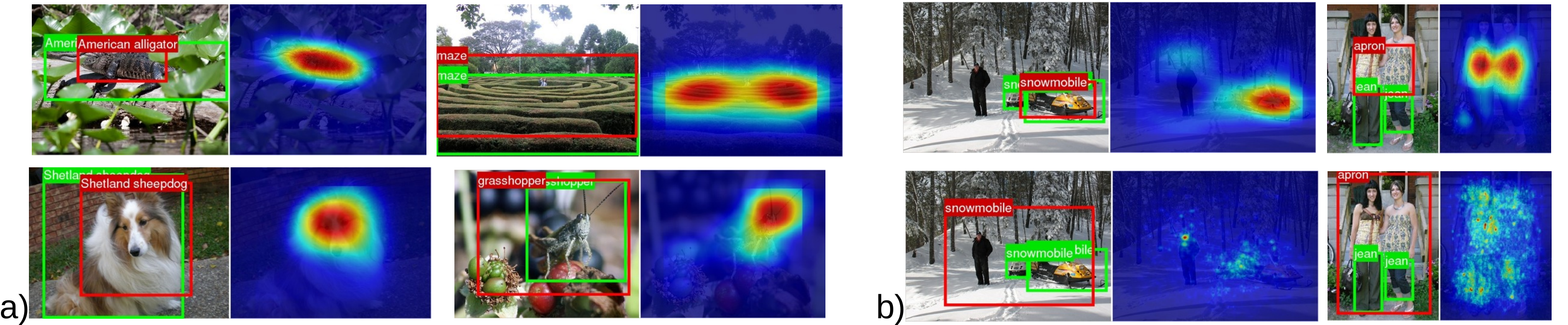}
\end{center}
\caption{a) Examples of localization from GoogleNet-GAP. b) Comparison of the localization from GooleNet-GAP (upper two) and the backpropagation using AlexNet (lower two). The ground-truth boxes are in green and the predicted bounding boxes from the class activation map are in red.}
\label{fig:localizationexample}
\end{figure*}

\textbf{Localization:} In order to perform localization, we need to generate a bounding box and its associated object category. To generate a bounding box from the CAMs, we use a simple thresholding technique to segment the heatmap. We first segment the regions of which the value is above 20\% of the max value of the CAM. Then we take the bounding box that covers the largest connected component in the segmentation map. We do this for each of the top-5 predicted classes for the top-5 localization evaluation metric. Fig.~\ref{fig:localizationexample}(a) shows some example bounding boxes generated using this technique. The localization performance on the ILSVRC validation set is shown in Tbl.~\ref{networkvalidationset}, and example outputs in Fig.~\ref{fig:activationmap}.

We observe that our GAP networks outperform all the baseline approaches with GoogLeNet-GAP achieving the lowest localization error of $43\%$ on top-5. This is remarkable given that this network was not trained on a single annotated bounding box. We observe that our CAM approach significantly outperforms the backpropagation approach of~\cite{simonyan2013deep} (see Fig.~\ref{fig:localizationexample}(b) for a comparison of the outputs). Further, we observe that GoogLeNet-GAP significantly outperforms GoogLeNet on localization, despite this being reversed for classification. We believe that the low mapping resolution of GoogLeNet ($7\times7$) prevents it from obtaining accurate localizations. Last, we observe that GoogLeNet-GAP outperforms GoogLeNet-GMP by a reasonable margin illustrating the importance of average pooling over max pooling for identifying the extent of objects.

\begin{table}\caption{Classification error on the ILSVRC validation set.}
\label{network_classificationValidation}
\centering
\footnotesize
\begin{tabular}{ l | c | c }
  \hline  
  \hline                       
  Networks & top-1 val. error & top-5 val. error \\
    \hline   
VGGnet-GAP& 33.4 & 12.2 \\
GoogLeNet-GAP& 35.0 & 13.2 \\
AlexNet$^{*}$-GAP & 44.9 & 20.9 \\
AlexNet-GAP & 51.1 & 26.3 \\
\hline
  GoogLeNet & 31.9 & 11.3 \\ 
  VGGnet & 31.2 &  11.4 \\    
  AlexNet &  42.6 &  19.5  \\
    NIN & 41.9 & 19.6 \\
    \hline 
    GoogLeNet-GMP & 35.6 & 13.9 \\    
    \hline
\end{tabular}
\end{table}

To further compare our approach with the existing weakly-supervised~\cite{simonyan2013deep} and fully-supervised~\cite{szegedy2014going,sermanet2013overfeat,szegedy2014going} CNN methods, we evaluate the performance of GoogLeNet-GAP on the ILSVRC test set. We follow a slightly different bounding box selection strategy here: we select two bounding boxes (one tight and one loose) from the class activation map of the top 1st and 2nd predicted classes and one loose bounding boxes from the top 3rd predicted class. We found that this heuristic was helpful to improve performances on the validation set. The performances are summarized in Tbl.~\ref{networktestset}. GoogLeNet-GAP with heuristics achieves a top-5 error rate of 37.1\% in a weakly-supervised setting, which is surprisingly close to the top-5 error rate of AlexNet (34.2\%) in a fully-supervised setting. While impressive, we still have a long way to go when comparing the fully-supervised networks with the same architecture (i.e., weakly-supervised GoogLeNet-GAP vs fully-supervised GoogLeNet) for the localization.

\begin{table}\caption{Localization error on the ILSVRC validation set. \textit{Backprop} refers to using~\cite{simonyan2013deep} for localization instead of CAM.}
\label{networkvalidationset}
\centering
\footnotesize
\begin{tabular}{ l | c | c }   
  \hline  
  \hline
  Method & top-1 val.error & top-5 val. error \\
    \hline  
  GoogLeNet-GAP&  \textbf{56.40} & \textbf{43.00} \\
  VGGnet-GAP& 57.20 & 45.14 \\
  GoogLeNet & 60.09 & 49.34\\      
  AlexNet$^{*}$-GAP & 63.75 & 49.53 \\     
  AlexNet-GAP & 67.19 & 52.16 \\
  NIN  & 65.47 & 54.19 \\
    \hline    
  Backprop on GoogLeNet & 61.31 & 50.55 \\  
  Backprop on VGGnet & 61.12 & 51.46 \\
  Backprop on AlexNet & 65.17 & 52.64 \\
  \hline  
    GoogLeNet-GMP  & 57.78 & 45.26 \\
    \hline 
\end{tabular}
\end{table}

\begin{table}\caption{Localization error on the ILSVRC test set for various weakly- and fully- supervised methods.}\label{networktestset}
\centering
\footnotesize
\begin{tabular}{ l | c | c  }
  \hline  
  \hline                       
  Method & supervision &  top-5 test error \\
    \hline  
  GoogLeNet-GAP (heuristics)  & weakly & \textbf{37.1} \\     
  GoogLeNet-GAP  & weakly & 42.9 \\ 
  Backprop \cite{simonyan2013deep} & weakly & 46.4 \\
      \hline 
  GoogLeNet \cite{szegedy2014going} & full & 26.7 \\
  OverFeat \cite{sermanet2013overfeat} & full & 29.9 \\        
  AlexNet \cite{szegedy2014going} & full & 34.2 \\
  \hline  
\end{tabular}
\end{table}

\section{Deep Features for Generic Localization}

The responses from the higher-level layers of CNN (e.g., \texttt{fc6}, \texttt{fc7} from AlexNet) have been shown to be very effective generic features with state-of-the-art performance on a variety of image datasets~\cite{donahue2014decaf,razavian2014cnn,zhou2014learning}. Here, we show that the features learned by our GAP CNNs also perform well as generic features, and as bonus, identify the discriminative image regions used for categorization, despite not having being trained for those particular tasks. To obtain the weights similar to the original softmax layer, we simply train a linear SVM~\cite{fan2008liblinear} on the output of the GAP layer.


First, we compare the performance of our approach and some baselines on the following scene and object classification benchmarks: SUN397~\cite{xiao2010sun}, MIT Indoor67~\cite{quattoni2009recognizing}, Scene15~\cite{lazebnik2006beyond}, SUN Attribute~\cite{patterson2012sun}, Caltech101~\cite{fei2007learning}, Caltech256~\cite{griffin2007caltech}, Stanford Action40~\cite{yao2011human}, and UIUC Event8~\cite{li2007and}. The experimental setup is the same as in~\cite{zhou2014learning}. In Tbl.~\ref{dataset_comparison}, we compare the performance of features from our best network, GoogLeNet-GAP, with the \texttt{fc7} features from AlexNet, and \texttt{ave pool} from GoogLeNet. 

As expected, GoogLeNet-GAP and GoogLeNet significantly outperform AlexNet. Also, we observe that GoogLeNet-GAP and GoogLeNet perform similarly despite the former having fewer convolutional layers. Overall, we find that GoogLeNet-GAP features are competitive with the state-of-the-art as generic visual features.

\begin{figure*}
\begin{center}
\includegraphics[width=1\textwidth]{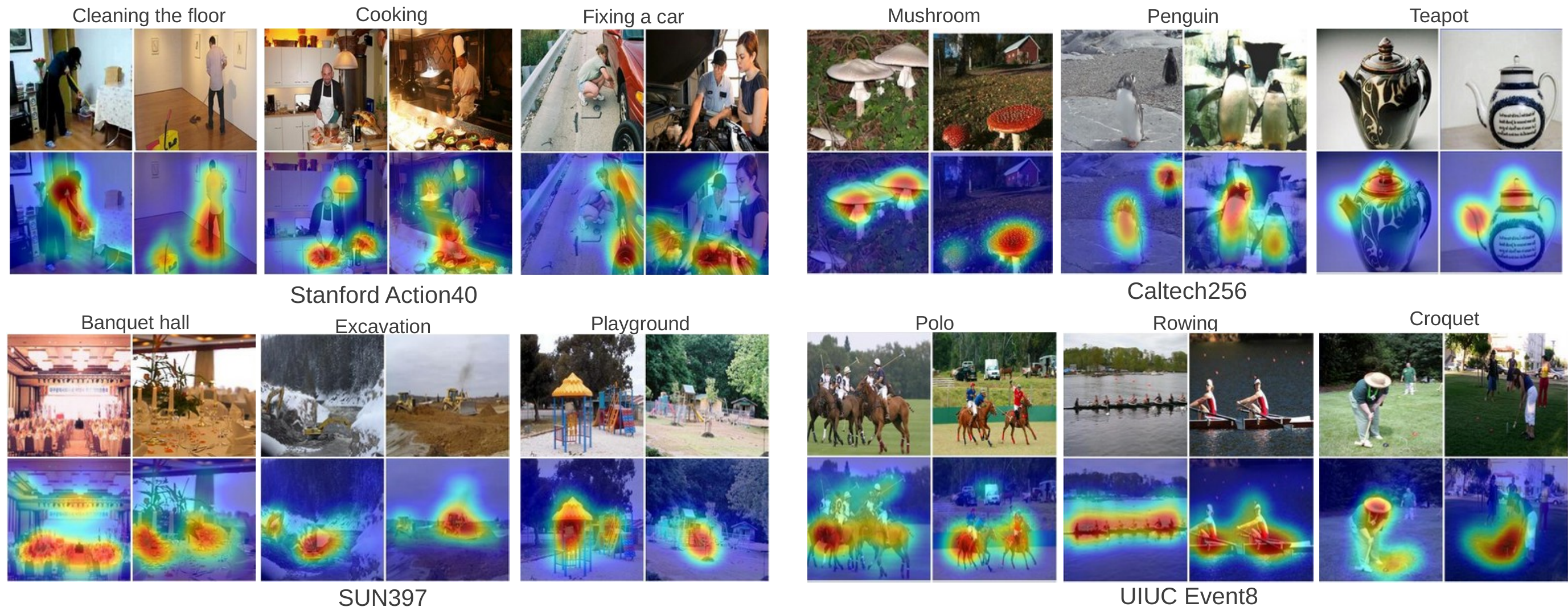}
\end{center}
\vspace*{-4mm}
\caption{Generic discriminative localization using our GoogLeNet-GAP deep features (which have been trained to recognize objects). We show 2 images each from 3 classes for 4 datasets, and their class activation maps below them. We observe that the discriminative regions of the images are often highlighted e.g., in Stanford Action40, the mop is localized for \textit{cleaning the floor}, while for \textit{cooking} the pan and bowl are localized and similar observations can be made in other datasets. This demonstrates the generic localization ability of our deep features.}
\label{fig:genericlocalization}
\end{figure*}

More importantly, we want to explore whether the localization maps generated using our CAM technique with GoogLeNet-GAP are informative even in this scenario. Fig.~\ref{fig:genericlocalization} shows some example maps for various datasets. We observe that the most discriminative regions tend to be highlighted across all datasets. Overall, our approach is effective for generating localizable deep features for generic tasks.

In Sec.~\ref{sec:finegrained}, we explore fine-grained recognition of birds and demonstrate how we evaluate the generic localization ability and use it to further improve performance. In Sec.~\ref{sec:pattern} we demonstrate how GoogLeNet-GAP can be used to identify generic visual patterns from images.

\subsection{Fine-grained Recognition}
\label{sec:finegrained}

In this section, we apply our generic localizable deep features to identifying 200 bird species in the CUB-200-2011~\cite{WelinderEtal2010} dataset. The dataset contains 11,788 images, with 5,994 images for training and 5,794 for test. We choose this dataset as it also contains bounding box annotations allowing us to evaluate our localization ability. Tbl.~\ref{birdresult} summarizes the results.

We find that GoogLeNet-GAP performs comparably to existing approaches, achieving an accuracy of 63.0\% when using the full image without any bounding box annotations for both train and test. When using bounding box annotations, this accuracy increases to 70.5\%. Now, given the localization ability of our network, we can use a similar approach as Sec.~\ref{sec:locresults} (i.e., thresholding) to first identify bird bounding boxes in both the train and test sets. We then use GoogLeNet-GAP to extract features again from the crops inside the bounding box, for training and testing. We find that this improves the performance considerably to 67.8\%. This localization ability is particularly important for fine-grained recognition as the distinctions between the categories are subtle and having a more focused image crop allows for better discrimination.

\begin{figure}
\begin{center}
\includegraphics[width=1\linewidth]{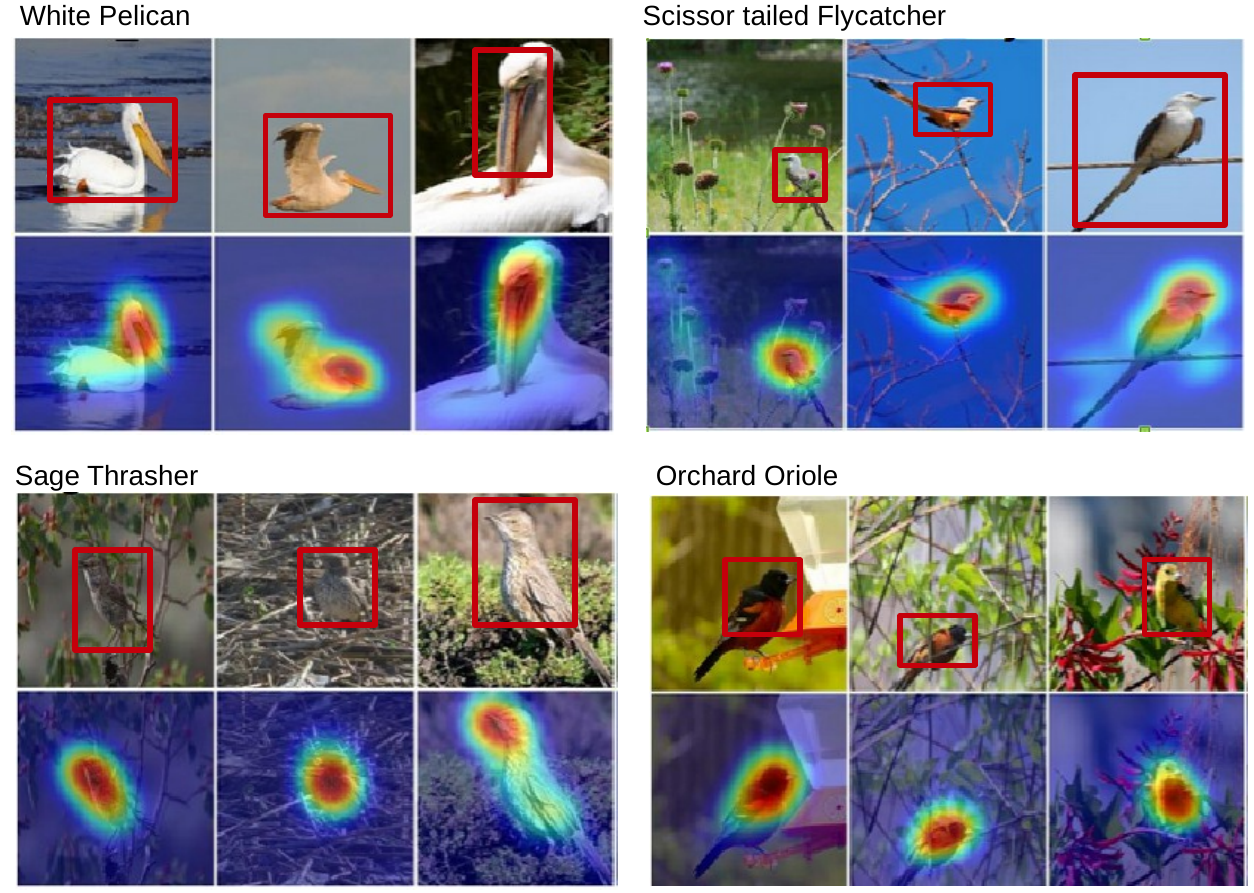}
\end{center}
\caption{CAMs and the inferred bounding boxes (in red) for selected images from four bird categories in CUB200. In Sec.~\ref{sec:finegrained} we quantitatively evaluate the quality of the bounding boxes (41.0\% accuracy for 0.5 IoU). We find that extracting GoogLeNet-GAP features in these CAM bounding boxes and re-training the SVM improves bird classification accuracy by about 5\% (Tbl.~\ref{birdresult}).}
\label{fig:bird}
\end{figure}

Further, we find that GoogLeNet-GAP is able to accurately localize the bird in 41.0\% of the images under the 0.5 intersection over union (IoU) criterion, as compared to a chance performance of 5.5\%. We visualize some examples in Fig.~\ref{fig:bird}. This further validates the localization ability of our approach.

\begin{table}\caption{Fine-grained classification performance on CUB200 dataset. GoogLeNet-GAP can successfully localize important image crops, boosting classification performance.}
\centering
\footnotesize
\begin{tabular}{ l | c | c }
\hline
  \hline                       
  Methods & Train/Test Anno. & Accuracy \\
    \hline  
    GoogLeNet-GAP on full image & n/a & 63.0\% \\
    GoogLeNet-GAP on crop & n/a & 67.8\% \\
	GoogLeNet-GAP on BBox & BBox & 70.5\% \\
    \hline   
    Alignments \cite{gavves2014local} & n/a & 53.6\% \\
    Alignments \cite{gavves2014local} & BBox & 67.0\%\\
    DPD \cite{zhang2013deformable} & BBox+Parts & 51.0\% \\
    DeCAF+DPD \cite{donahue2014decaf} & BBox+Parts & 65.0\%\\
    PANDA R-CNN \cite{zhang2014part} & BBox+Parts & 76.4\% \\
    \hline  
\end{tabular}\label{birdresult}
\end{table}

\begin{table*}\caption{Classification accuracy on representative scene and object datasets for different deep features. }\label{dataset_comparison}
\centering
\footnotesize
\begin{tabular}{lcccccccc}
\hline
\hline
 &SUN397&MIT Indoor67&Scene15&SUN Attribute&Caltech101&Caltech256&Action40 & Event8 \\
\hline
\texttt{fc7} from AlexNet       & 42.61 & 56.79 & 84.23 & 84.23 & 87.22 & 67.23 & 54.92 & 94.42 \\
\texttt{ave pool} from GoogLeNet     & 51.68 & 66.63 & 88.02 & 92.85 & 92.05  & 78.99    & 72.03 & 95.42\\
\texttt{gap} from GoogLeNet-GAP  & 51.31 & 66.61 & 88.30 & 92.21 & 91.98  & 78.07    & 70.62 & 95.00\\
\hline
\end{tabular}
\label{tableResultsDeepSceneFeat}
\end{table*}

\subsection{Pattern Discovery}
\label{sec:pattern}

In this section, we explore whether our technique can identify common elements or patterns in images beyond objects, such as text or high-level concepts. Given a set of images containing a common concept, we want to identify which regions our network recognizes as being important and if this corresponds to the input pattern. We follow a similar approach as before: we train a linear SVM on the GAP layer of the GoogLeNet-GAP network and apply the CAM technique to identify important regions. We conducted three pattern discovery experiments using our deep features. The results are summarized below. Note that in this case, we do not have train and test splits $-$ we just use our CNN for visual pattern discovery.



\begin{figure}
\begin{center}
\includegraphics[width=1\linewidth]{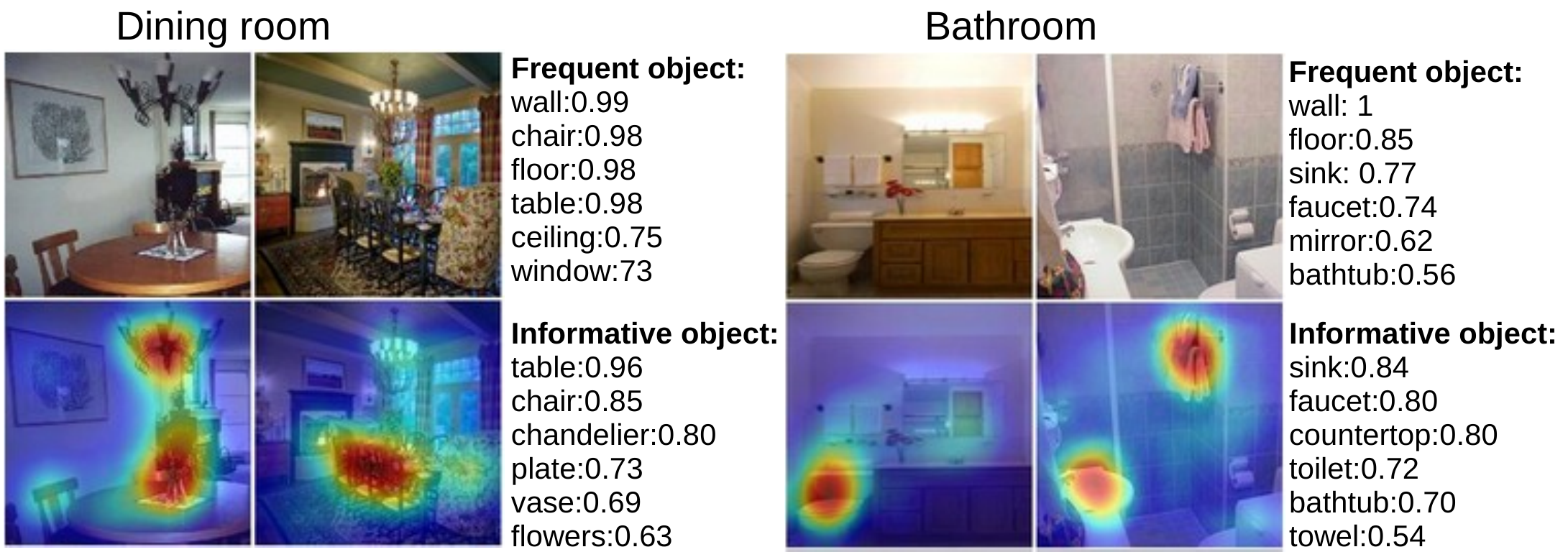}
\end{center}
\vspace*{-4mm}
\caption{Informative objects for two scene categories. For the dining room and bathroom categories, we show examples of original images (top), and list of the $6$ most frequent objects in that scene category with the corresponding frequency of appearance. At the bottom: the CAMs and a list of the 6 objects that most frequently overlap with the high activation regions.}
\label{fig:scene}
\end{figure}

\textbf{Discovering informative objects in the scenes:} We take 10 scene categories from the SUN dataset~\cite{xiao2010sun} containing at least $200$ fully annotated images, resulting in a total of 4675 fully annotated images. We train a one-vs-all linear SVM for each scene category and compute the CAMs using the weights of the linear SVM. In Fig.~\ref{fig:scene} we plot the CAM for the predicted scene category and list the top 6 objects that most frequently overlap with the high CAM activation regions for two scene categories. We observe that the high activation regions frequently correspond to objects indicative of the particular scene category.

\begin{figure}
\begin{center}
\includegraphics[width=1\linewidth]{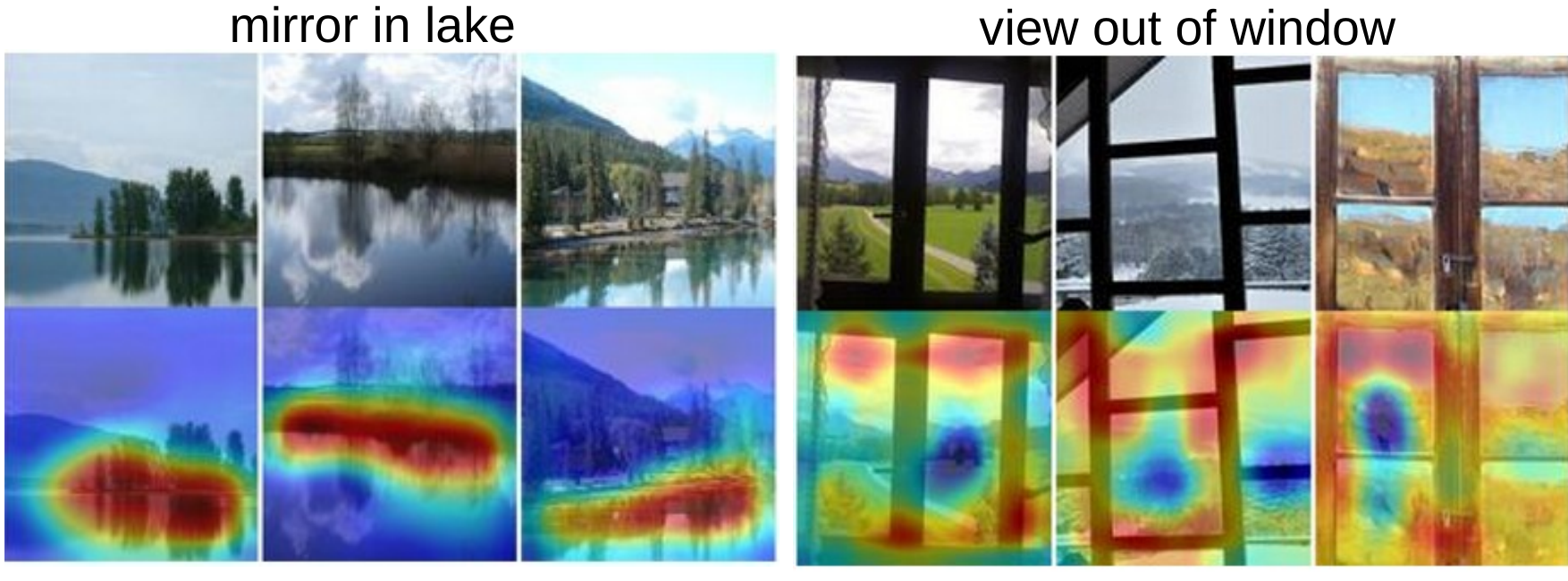}
\end{center}
\vspace*{-4mm}
\caption{Informative regions for the concept learned from weakly labeled images. Despite being fairly abstract, the concepts are adequately localized by our GoogLeNet-GAP network.}
\label{fig:conceptlearner}
\end{figure}

\textbf{Concept localization in weakly labeled images:} Using the hard-negative mining algorithm from~\cite{zhou2014conceptlearner}, we learn concept detectors and apply our CAM technique to localize  concepts in the image. To train a concept detector for a short phrase, the positive set consists of  images that contain the short phrase in their text caption, and the negative set is composed of randomly selected images without any relevant words in their text caption. In Fig.~\ref{fig:conceptlearner}, we visualize the top ranked images and CAMs for two concept detectors. Note that CAM localizes the informative regions for the concepts, even though the phrases are much more abstract than typical object names.

\begin{figure}
\begin{center}
\includegraphics[width=1\linewidth]{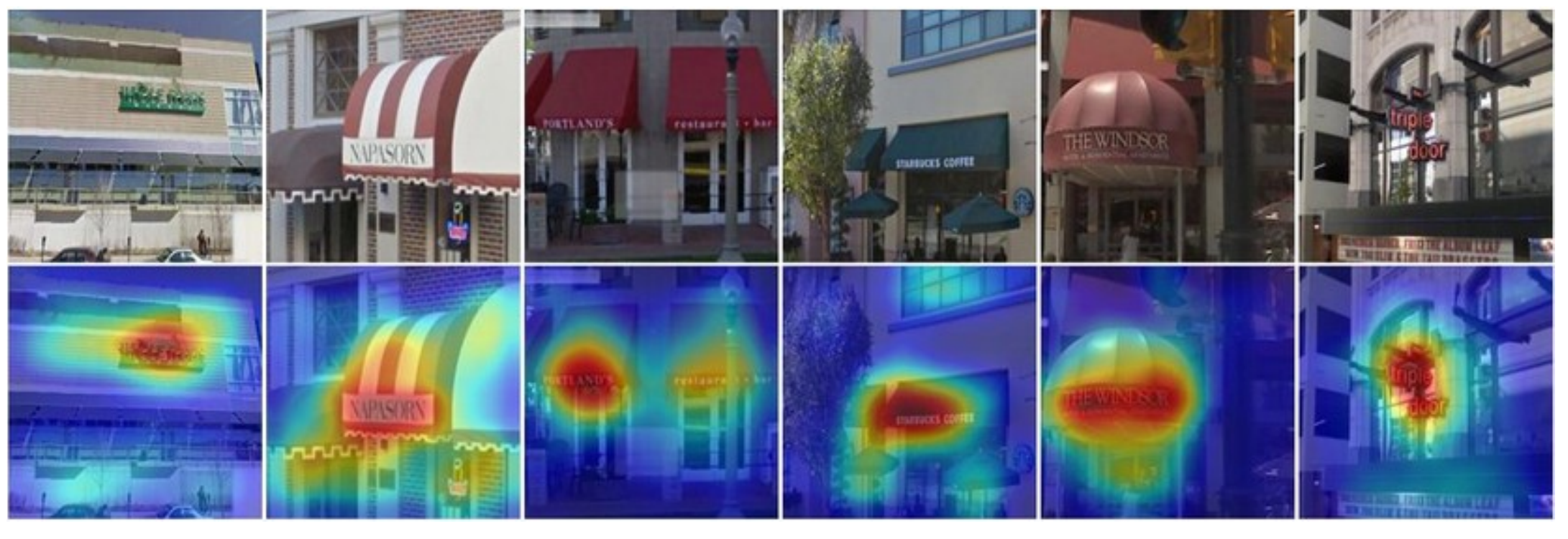}
\end{center}
\vspace*{-4mm}
\caption{Learning a weakly supervised text detector. The text is accurately detected on the image even though our network is not trained with  text or any bounding box annotations.}
\label{fig:textdetection}
\end{figure}

\textbf{Weakly supervised text detector:} We train a weakly supervised text detector using 350 Google StreetView images containing text from the SVT dataset~\cite{wang2011end} as the positive set and randomly sampled images from outdoor scene images in the SUN dataset \cite{xiao2010sun} as the negative set. As shown in Fig.~\ref{fig:textdetection}, our approach highlights the text accurately without using bounding box annotations.

\begin{figure}
\begin{center}
\includegraphics[width=1\linewidth]{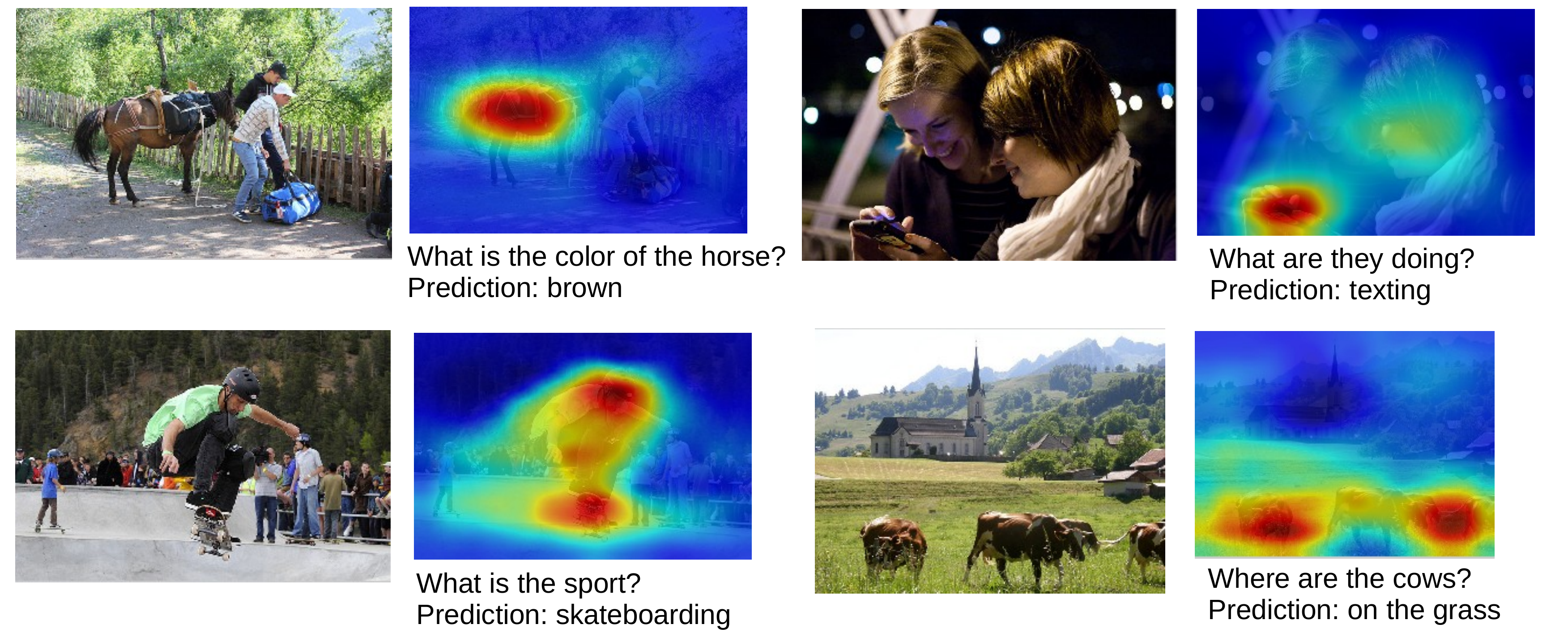}
\end{center}
\caption{Examples of highlighted image regions for the predicted answer class in the visual question answering.}
\label{fig:vqa}
\end{figure}

 \textbf{Interpreting visual question answering:} We use our approach and localizable deep feature in the baseline proposed in \cite{zhou2015vqa} for visual question answering. It has overall accuracy 55.89\% on the test-standard in the Open-Ended track. As shown in Fig.~\ref{fig:vqa}, our approach highlights the image regions relevant to the predicted answers.

\section{Visualizing Class-Specific Units}

Zhou \textit{et al}~\cite{zhou2014object} have shown that the convolutional units of various layers of CNNs act as visual concept detectors, identifying  low-level concepts like textures or materials, to high-level concepts like objects or scenes. Deeper into the network, the units become increasingly discriminative. However, given the fully-connected layers in many networks, it can be difficult to identify the importance of different units for identifying different categories. Here, using GAP and the ranked softmax weight, we can directly visualize the units that are most discriminative for a given class. Here we call them the \textit{class-specific units} of a CNN. 

\begin{figure}
\begin{center}
 \includegraphics[width=1\linewidth]{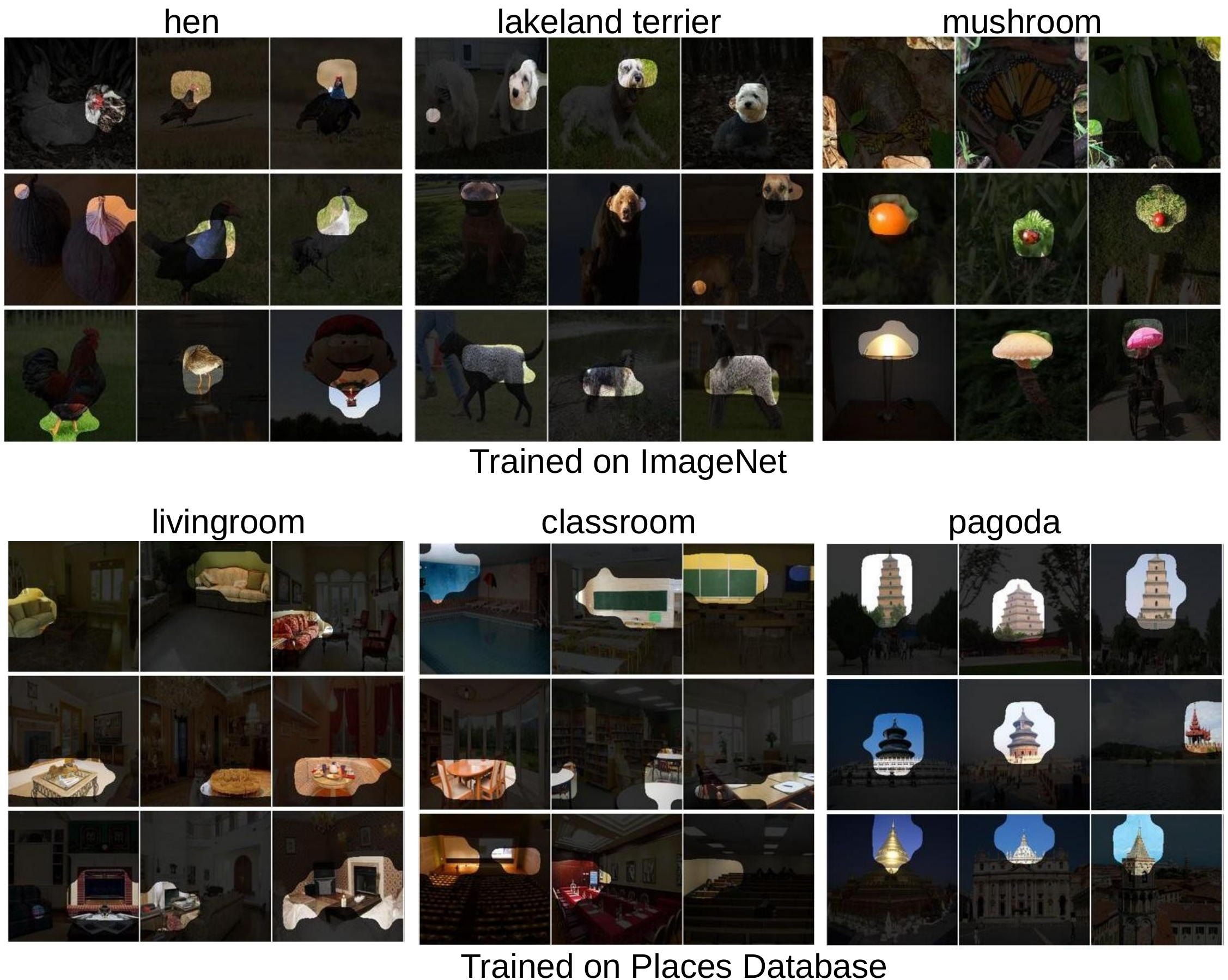}
\end{center}
\vspace*{-4mm}
\caption{Visualization of the class-specific units for AlexNet*-GAP trained on ImageNet (top) and Places (bottom) respectively. The top 3 units for three selected classes are shown for each dataset. Each row shows the most confident images segmented by the receptive field of that unit. For example, units detecting blackboard, chairs, and tables are important to the classification of  \textit{classroom} for the network trained for scene recognition.}
\label{fig:unitvisualization}
\end{figure}

Fig.~\ref{fig:unitvisualization} shows the class-specific units for AlexNet$^{*}$-GAP trained on ILSVRC dataset for object recognition (top) and Places Database for scene recognition (bottom).
We follow a similar procedure as~\cite{zhou2014object} for estimating the receptive field and segmenting the top activation images of each unit in the final convolutional layer. Then we simply use the softmax weights to rank the units for a given class. From the figure we can identify the parts of the object that are most discriminative for classification and exactly which units detect these parts. For example, the units detecting dog face and body fur are important to  \textit{lakeland terrier}; the units detecting sofa, table and fireplace are important to the  \textit{living room}. Thus we could infer that the CNN actually learns a bag of words, where each word is a discriminative class-specific unit. A combination of these class-specific units guides the CNN in classifying each image.

\section{Conclusion}


In this work we propose a general technique called Class Activation Mapping (CAM) for CNNs with global average pooling. This enables classification-trained CNNs to learn to perform object localization, without using any bounding box annotations. Class activation maps allow us to visualize the predicted class scores on any given image, highlighting the discriminative object parts detected by the CNN. We evaluate our approach on weakly supervised object localization on the ILSVRC benchmark, demonstrating that our global average pooling CNNs can perform accurate object localization. Furthermore we demonstrate that the CAM localization technique generalizes to other visual recognition tasks i.e., our technique produces generic localizable deep features that can aid other researchers in understanding the basis of discrimination used by CNNs for their tasks.

{\small
\bibliographystyle{ieee}
\bibliography{egbib}
}

\end{document}